\documentclass[twocolumn]{wlscirep}
\usepackage{inputenc}
\usepackage{amsmath,amssymb, amsthm}
\usepackage{subcaption}
\usepackage{cleveref}
\usepackage{graphicx}
\usepackage{bbm}
\usepackage{color}
\usepackage{comment}
\usepackage{amsthm}
\usepackage[normalem]{ulem}
\usepackage{bm}
\usepackage[textsize=tiny]{todonotes}

\newtheorem{definition}{Definition}
\newif\iflong
\longtrue

\newcommand{\Hy}{\mathcal{H}}
\newcommand{\model}{HyDy-GNN }

\newcommand{\multiset}[1]{\ensuremath{\{\mkern-7mu\{ #1 \}\mkern-7mu\}}}

\title{Learning the effective order of a hypergraph dynamical system}
\author[1]{Leonie Neuhäuser}
\author[1]{Michael Scholkemper}
\author[2]{Francesco Tudisco}
\author[1]{Michael T. Schaub}

\affil[1]{RWTH Aachen University, Aachen, Germany}
\affil[2]{GSSI Gran Sasso Science Institute, L’Aquila, Italy}
\begin{abstract}
    Dynamical systems on hypergraphs can display a rich set of behaviours not observable for systems with pairwise interactions.
    Given a distributed dynamical system with a putative hypergraph structure, an interesting question is thus how much of this hypergraph structure is actually necessary to faithfully replicate the observed dynamical behaviour.
    To answer this question, we propose a method to determine the minimum order of a hypergraph necessary to approximate the corresponding dynamics accurately. 
    Specifically, we develop an analytical framework that allows us to determine this order when the type of dynamics is known. 
    We utilize these ideas in conjunction with a hypergraph neural network to directly learn the dynamics itself and the resulting order of the hypergraph from both synthetic and real data sets consisting of observed system trajectories. 
\end{abstract}

\begin{document}

\maketitle
Dynamical processes on hypergraphs~\cite{majhi_dynamics_2022,battiston_networks_2020} have recently received significant attention.
Examples include synchronisation~\cite{lucas_multiorder_2020,skardal_higher_2020}, consensus dynamics~\cite{neuhauser_multibody_2020,neuhauser_opinion_2021,sahasrabuddhe_modelling_2020}, epidemic spread~\cite{bodo_sis_2016,landry_effect_2020,higham_epidemics_2021}, random walks~\cite{carletti_random_2020,chitra_random_2019}, label propagation~\cite{liu2021strongly,prokopchik2022nonlinear,fountoulakis2021local} and social contagion~\cite{iacopini_simplicial_2019,de_arruda_social_2020}. 
Generally, it has been found that for dynamical systems supported on hypergraphs instead of graphs, important characteristics of the dynamics can change. 
For example, contagion processes on hypergraphs can have different epidemic thresholds for the outbreak of an epidemic~\cite{higham_epidemics_2021, landry_effect_2020}, or group reinforcement effects can significantly alter the outcome of opinion formation on hypergraphs~\cite{neuhauser_consensus_2022,hickok_bounded-confidence_2022,de_arruda_social_2020}.
Hypergraphs have also gained interest as extensions of graph neural networks and been used for a variety of applications ranging from 3D shape retrieval~\cite{bai_multi-scale_2021} and pose estimation~\cite{liu_semi-dynamic_2020} in computer vision, to group recommendation tasks~\cite{guo_hierarchical_2021}, or medical applications such as cancer tissue classification~\cite{bakht_colorectal_2021}. 

Although both aspects often appear lumped together in the mathematical equations describing a dynamical system on a hypergraph, from a modeling perspective, it is important to distinguish between (i) the topological relations constraining the possible interactions in a system (as encoded in a hypergraph) and (ii) the model of the local multi-way dynamics occurring on each hyperedge (e.g., epidemic spread, diffusion, synchronisation).
It is the interplay of both aspects that leads to the (possible) emergence of higher-order effects~\cite{bick2021higher,battiston_networks_2020,neuhauser_consensus_2022}.

For instance, if we consider a linear dynamics on a hypergraph, we cannot expect any higher-order effects to emerge.
As any linear map between finite-dimensional spaces can be represented by a matrix, for linear dynamics we may always find an equivalent graph-based, pairwise interaction dynamics that models the system exactly, by identifying the matrix with an effective weighted graph~\cite{neuhauser_multibody_2020,neuhauser_opinion_2021}.
In practice, this means that it is possible to use a graph instead of a hypergraph-based dynamical system, as long as linear dynamics are considered. 
Similarly, it has been shown that various formulations of semi-supervised and unsupervised spectral clustering problems on hypergraphs lead to an effective graph-theoretic problem~\cite{agarwal_higher_2006,veldt2022hypergraph,tudisco2023core}. 

More generally, we can envision that, depending on the local dynamics, we will be able to rewrite a hypergraph dynamical system on a hypergraph of general order $k$ as a dynamics occurring on a hypergraph with hyperedges of order at most $p\leq k$.
Such a simplification to lower-order relations is relevant as the use of hypergraphs presents several challenges: most prominently, since the number of hyperedges can grow combinatorially with the number of nodes, the use of hypergraph models can be computationally very expensive. 
This is particularly relevant for large-scale systems. 

Yet, in practice, we typically know neither the exact set of relations between the entities, nor the analytical form of the local interaction dynamics, but are only given observational data, e.g., in the form of trajectories. 
Many different approaches have been explored to approximate such time series data. 
For example, it has recently been proposed to treat neural networks as models equipped with a continuum of layers~\cite{chen_neural_2018}. 
This view allows a reformulation of the forward pass of a neural network as the solution of an initial value problem of an ordinary differential equation (ODE).
Deep learning architectures that use this reinterpretation of the forward pass are called Neural ODEs and are useful for building continuous-time time series models. 
Neural ODEs have also been recently generalised to graphs~\cite{zang_neural_2020,poli_graph_2021}.
Discovering the equations governing a dynamical system from measurements is an important problem, tackled by a large body of literature~\cite{brunton_discovering_2016,mangan_inferring_2016,bongard_automated_2007,schmidt_distilling_2009}. 
The main challenge here is to find a model which is complex enough to describe the existing data but not too complex to avoid overfitting. 

For a dynamical system on a hypergraph, there are many possible ways to abstract an observed distributed dynamics.
For instance, we may model it as emerging from a simple local dynamics on a rather complex hypergraph.
Alternatively, we may consider a more complicated local dynamics interacting via a more constrained set of relations between entities. 
This raises the question of whether we should include the complexity of our model in the topology of the interactions or in the model of the dynamics:
what multi-way relations do we need to encode in our model in practice?

In this paper, we thus introduce the concept of the dynamical order of a hypergraph dynamical system, which measures how complex the dynamics are. 
Based on this dynamical order, we can determine the effective order of the hypergraph dynamical system, which is given by the minimum order of a hypergraph necessary to accurately represent the corresponding dynamics. 
In particular, we propose an analytical framework that allows us to derive the dynamical and the effective order when the functional form of the dynamics is given. 
Furthermore, we propose a hypergraph neural network architecture that allows learning the hypergraph dynamics and the resulting effective order directly from data, which we test on both synthetic and real data sets. 
In conclusion, we present an effective method to reduce the complexity of a hypergraph dynamical system and to learn its representation from data.

\section{Reducibility of hypergraph dynamical systems}

To illustrate our ideas of the dynamical and effective order of a dynamical system on a hypergraph, let us start with a Kuramoto-type dynamics~\cite{kuramoto_chemical_2003} on a hypergraph as a concrete example. Kuramoto oscillator dynamics have been applied to various synchronization phenomena of phase oscillators~\cite{Stankovski-2017}, ranging from power networks~\cite{dorfler_synchronization_2010} to brain activity~\cite{cabral_exploring_2014}. Several works have been working on its generalisation to simplicial complexes~\cite{skardal_higher_2020,lucas_multiorder_2020} and hypergraphs~\cite{adhikari_synchronization_2023}. We will compare two different formulations of Kuramoto dynamics on a hypergraph here.

Consider a hypergraph $\Hy$ consisting of a set ${\mathcal{V} = \{ 1, 2, \hdots, N \}}$ of $N$ nodes, and a set ${\mathcal{E} = \{ E_1, E_2, \hdots, E_M \}}$ of $M$ hyperedges.
Each hyperedge $E_\alpha$ is a subset of the nodes, i.e., $E_\alpha \subseteq \mathcal{V}$ for all $\alpha =1,2,\hdots,M$. 
Each hyperedge may have a different cardinality $|E_\alpha|$ . 
We define the topological order $k=\max_\alpha|E_\alpha|$ as the cardinality of the largest hyperedge.
Let $x(t) \in \mathbb{R}^N$ be the vector of dynamical state variables of the nodes at time $t$. 
For simplicity, we will suppress the time dependency and write $x_i=x_i(t)$ for the $i$th component of the node state vector in the following.

There are several ways in which the well-known Kuramoto dynamics can be extended to hypergraphs.
First, inspired by Adhikari et al. \cite{adhikari_synchronization_2023} we can write the time evolution of node $i$ as 
\begin{align}\label{eq:Kuramoto_higher}
\dot{x_i} = \sum_{E_\alpha: i \in E_\alpha} \sin\left( \sum_{j \in E_\alpha} (x_j- x_i)\right).
\end{align}
Since the non-linear sine function acts within each entire hyperedge, for general node states $x_i$ it is not clear how to further reduce the system to a dynamical system on a lower-order hypergraph.

An alternative formulation for Kuramoto oscillator dynamics is
\begin{align}\label{eq:Kuramoto_pairwise}
  \dot{x_i} = \sum_{\alpha: i \in E_\alpha} \sum_{j \in E_\alpha} \sin(x_j- x_i) = \sum_{j=1}^{N} (B^\top B)_{ij} \sin(x_j - x_i)
  \end{align}
  where within each hyperedge, every pair of nodes interacts. Here, $B \in \mathbb{R}^{M \times N}$ represents the node-to-edge incidence matrix
 \begin{align}
   B_{\alpha i } = \begin{cases}1 &\text{if } i \in E_\alpha, \\
   0 &\text{else.}
   \end{cases}
   \end{align}
Hence, even though we are dealing here with a nonlinear dynamics on a hypergraph, the dynamics in each hyperedge is pairwise and we can reduce it to a pairwise network dynamics: the hypergraph topology simply scales the system by $A=B^\top B$. 

Overall, our example of Kuramoto oscillator dynamics shows that whether the hypergraph can be projected onto a lower-order system depends on the form of the dynamics supported on each hyperedge: the dynamics is reducible if the dynamics on the hyperedges can be rewritten as a linear combination of lower-order functions. 
More generally, for certain functional forms the non-linear dynamics can always be reduced to a lower-order hypergraph system. 
In the Supplementary Materials in Section 1, we show the linear-like properties that dynamics must have in order to be reducible to a network dynamical system.

In the following, we formalise this form of dynamical reduction by distinguishing between the \emph{topological order} of a hypergraph and the \emph{dynamical order} of a dynamics. 
Combining topology and dynamics to a hypergraph dynamical system, its \emph{effective order} is given by the minimum of these two orders. 

\subsection*{Dynamical order and effective order}
\begin{figure*}
  \centering
  \begin{subfigure}{\textwidth}
    \centering
    \includegraphics[width=\textwidth]{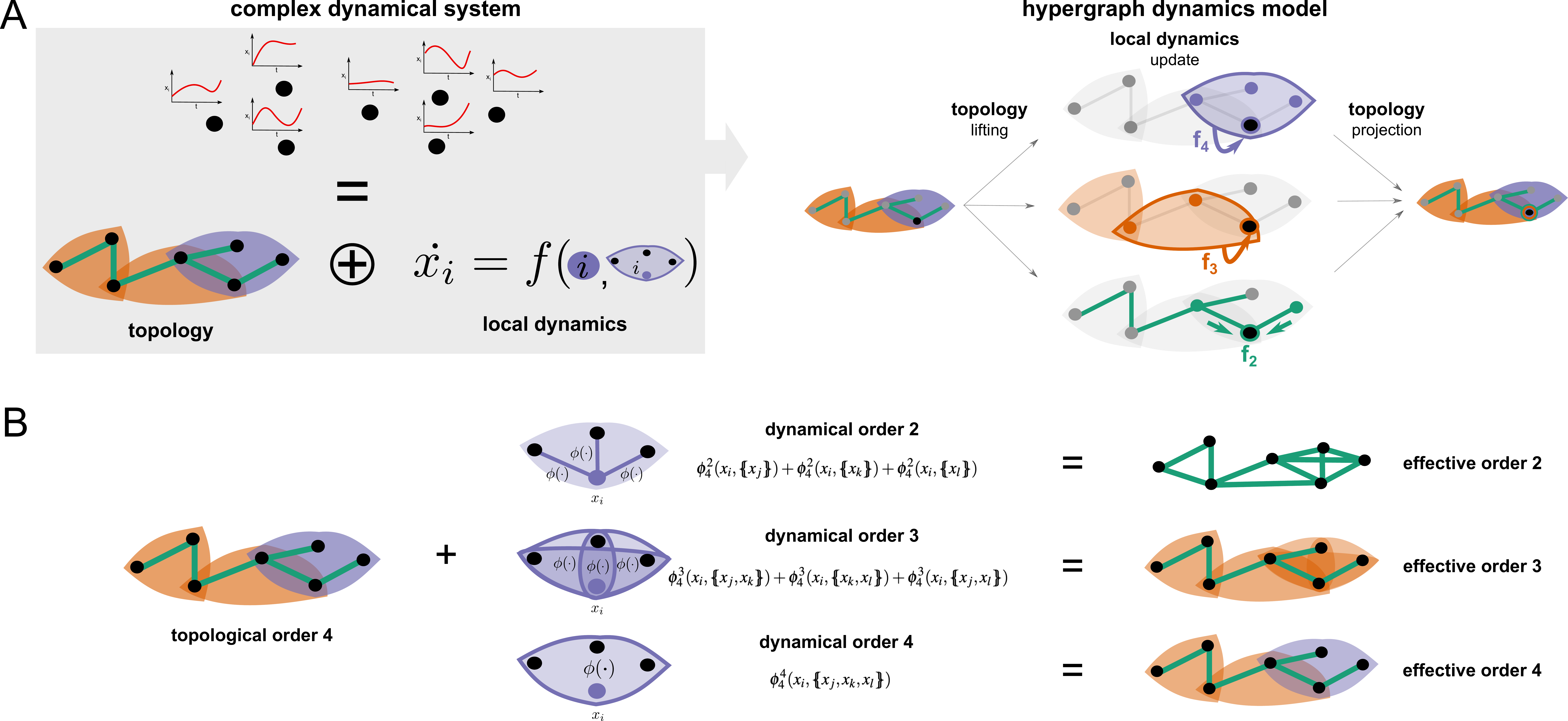}
  \end{subfigure}
  \caption[Hypergraph dynamics model and visualisation of the topological, dynamical and effective order.]{\textbf{Hypergraph dynamics model and visualisation of the topological, dynamical and effective order.} When modelling complex dynamical systems, it is necessary to take into account both the system's topology and the local dynamics which are acting on the topology. We introduce a hypergraph dynamics model which separates the impact of topology and dynamics. 
  In A, we show the update process of a single node according to the full hypergraph topology. All nodes in the hypergraph are updated synchronously. In particular, the values of the nodes that are part of hyperedges of size $d$ are collected. Then an update function on these hyperedges which defines the dynamics is computed to obtain the update and then projected back to the node space by summing over each update. With the help of this framework, we can define and analyse different orders of the system, which we visualise in B. The topological order is given by the size of the largest hyperedge in the hypergraph. To derive the dynamical order, we look at the update functions on a hyperedge of topological order $4$ in more detail. In particular, the update resulting from a hyperedge of topological order $4$ can consist of a linear combination of pairwise or three-way functions which results in dynamical order $2$ or $3$, respectively. We can then derive the effective order of the hypergraph dynamics which is the minimum of the topological and dynamical order. The illustration highlights that the topological order of a hypergraph is only an upper bound, not a lower bound, on the effective order of the system.}
  \label{Fig1_concept}
\end{figure*}

Here we introduce a framework to separate the impact of topology and dynamics for a general hypergraph dynamical system. 
To this end, we group the edges of a hypergraph $\Hy$ by size, so that its edge set is given by $\mathcal{E}=\mathcal{E}_{2} \cup \dots \cup \mathcal{E}_{k}$ where $\mathcal{E}_{d}$ denotes the set of all hyperedges of size $d$. 
We refer to the values in the state vector $x(t)$ of nodes in the hyperedge $E$ at time $t$ excluding node $i$ with $\multiset{x_j(t) \mid j \in E, j \neq i}$. 
This is a multiset, which we denote by double braces --- i.e.\ a set where elements may occur multiple times --- since the node may have the same values. 

Formally, we now define a hypergraph dynamical system as follows.
\begin{definition}[Hypergraph dynamical system]
Consider a family of multivariate update functions $(f_d)_{d \in 1\ldots K}$ where each function $f_d(y_1,\dots,y_d)$ acts on $d$ variables, and $f_d$ is symmetric in its last $d-1$ arguments, i.e., any permutation the last $d-1$ variable does not change the function.
Together with the topology of a hypergraph $\Hy$ with topological order $k\leq K$, these update functions $f_d$ describe the local dynamics on each hyperedge.
The dynamics on a node can then be written in the form of:
\begin{align}\label{eq:hypergraph_dynamics}
\dot{x_i} = f_1(x_i)+\underbrace{\sum_{d=2}^k \; \sum_{ E \in \mathcal{E}_d: i \in E}}_{\text{topology} } \underbrace{f_d (x_i, \multiset{x_j \mid j \in E, j \neq i})}_{\text{dynamics}},
\end{align}
These update equations for $i=1\ldots N$ define a \emph{hypergraph dynamical system} on $\Hy$.
\end{definition}

Note that \Cref{eq:hypergraph_dynamics} separates the hypergraph topology from the dynamics on the edges: For each size $d$, a set of hyperedges defines the topological coupling of entities within the system and the update functions $f_d$ act on the $d$ state variables associated to the nodes within each of these hyperedges. 
This is visualised in \Cref{Fig1_concept} A. 
Details of the mathematical framework can be found in \Cref{sec:hypergraphlearningmodel}.
As we are primary interested in interacting dynamics, we will for simplicy set $f_1 =0$ in the following, even though all our arguments can be naturally extended to the nonzero case.

Most commonly used hypergraph dynamical processes considered in the recent literature can be written in the above form.
For example, the dynamics of the Kuramoto oscillator model on hyperedges in \Cref{eq:Kuramoto_higher} can be written as above if we define $f_1=0$ and: 
$$
f_d(y_1, \dots, y_d) = \sin\left(\sum_{i=1}^d(y_i-y_1)\right) \quad \text{for} \quad d=2,\dots, k.
$$

We  now introduce the \emph{dynamical order} of such a hypergraph dynamical system, determined by the form of the update functions $f_d$.
\begin{definition}[Dynamical order]\label{def:dynamical_order}
    Consider a hypergraph dynamical system.
    Each update function $f_d$ can be generically decomposed into a sum of functions $\phi_d^p$ of only $p\leq d$ variables:
  \begin{align}\label{eq:dynamical_order_function}
     f_d(y_1,\mathfrak{s}) = \sum_{\mathfrak{v} \subseteq \mathfrak{s} : |\mathfrak{v}|= p-1}\phi_d^{p}(y_1,\mathfrak{v}),
  \end{align} 
    where $\mathfrak{s}$ denotes a multi-set of variables $\multiset{y_2,\dots,y_d}$ of size $d-1$ and $\mathfrak{v}$ denotes a non-empty subset of $\mathfrak{s}$ of dimension $p-1$.
    We call the minimal $p\ge 2$\footnote{As are only concerned with update functions that have at least two function arguments (at least pairwise interactions) the dynamical order has to be $p_\text{dyn}\ge 2$.}
    for which this decomposition is possible for \emph{all} update functions $f_d$ the \emph{dynamical order} $p_{\text{dyn}}$ of the system.
\end{definition}
Trivially, by choosing $p=d$ we have that $\phi^d_d(y_i,\mathfrak{s}) = f_d(y_i,\mathfrak{s})$ such that the above decomposition always exists.
However, for specific functional forms $f_d$, it can be possible to decompose higher-order interactions into a combination of $p$-ary interactions. 
Let us illustrate this with a concrete example. 
Consider the family of update functions defined for all $d$ via $f_d(y_1,\dots,y_d) = \log(y_1\dots y_d)$, i.e., $f_1: y_1\mapsto \log(y_1)$, $f_2: (y_1,y_2)\mapsto \log(y_1y_2)$, and so on. 
By the properties of the logarithm we can write $f_d$ for any $d\ge 2$ as:
\begin{equation}
    f_d(y_1,\dots,y_d) =\sum_{i=2}^d \left(\frac{1}{d-1}\log(y_1)+\log(y_i)\right)= \sum_{i=2}^d\phi_d^2(y_1,y_i).
\end{equation}
and thus the dynamical order is $p_\text{dyn}=2$.

From our above discussion, it should be apparent that for a dynamical order $p_{\text{dyn}}$, the dynamics on the larger hyperedges always consist of linear combinations of functions of order $p_{\text{dyn}}$.
Hence, one may rewrite \Cref{eq:hypergraph_dynamics} more compactly in terms of dynamics on (effective) hyperedges of size $p_{\text{dyn}}$. 
This is akin to a higher-order equivalent of a clique expansion.~\cite{sun_hypergraph_2008}.

This leads us to the definition of the \emph{effective order}, which combines topological and dynamical order (see \cref{Fig1_concept}B).
\begin{definition}[Effective order]\label{def:effective_order}
  Consider a hypergraph dynamical system which consists of a hypergraph $\Hy$ with topological order $k$ and a dynamical system $(f_d)_{d \in \mathbb{N}}$ with dynamical order $p_{\text{dyn}}$. 
  The effective order $p_\text{min}$ of the system is the minimum order of a hypergraph necessary to describe the corresponding dynamics accurately. 
  It is bounded by
  \begin{align}
    p_{\text{min}}\leq\text{min}(k,p_{\text{dyn}}).
  \end{align}
\end{definition}

Indeed, if $k = p_{\text{dyn}}$, the dynamics on none of the hyperedges decompose and the effective order of the system is given by the topological order. Consequently, we cannot reduce the hypergraph dynamical system any further. This was the case in \cref{eq:Kuramoto_higher}. 
Thus, the effective order of this hypergraph Kuramoto model is $p_{\text{min}}=k$.

In contrast, if $p_{\text{dyn}}<k$, the dynamics on the hyperedges larger than the dynamical order factorise into $p_{\text{dyn}}$-ary functions, and the hypergraph dynamical system can be projected onto a (new, effective) hypergraph of order $p_{\text{min}}=p_{\text{dyn}}<k$
An example for an effective order of $p_{\text{min}}=2$ is given by \cref{eq:Kuramoto_pairwise} where the update functions for $d=2,\dots,k$ are:
\begin{align}
    f_d(y_1,\mathfrak{s})=\sum_{y_i \in \mathfrak{s}} \sin(y_1- y_i) = \sum_{y_i \in \mathfrak{s}}\phi_d^2(y_1,y_i). 
\end{align}
This dynamics has a dynamical order of $p_{\text{dyn}}=2<k$ and the dynamics thus factorise into pairwise functions on all hyperedges so that they are effectively network dynamics. 
Overall, the exchange of the order of the sum and the $\sin$ function is key to make the dynamical order of \cref{eq:Kuramoto_higher} and \cref{eq:Kuramoto_pairwise} differ, leading to different effective order of the systems.

Though conceptually useful, the derivation of the effective order as described in this section requires knowledge of the analytical form of the dynamics. 
In the next section, we thus present a method to learn the corresponding functions by deriving a hypergraph neural network model. 
Using this computational model, we can learn the dynamics from data --- and thus implicitly the effective order of the system --- without prior knowledge on the functional form of the dynamics.

\section{Learning local dynamics and effective order from data}
In this section, we present a method to learn the local dynamics and thus the dynamical and effective order of the system directly from data. 

\subsection{Learning a hypergraph dynamical system}
\label{ssec:learning}
In our formulation of hypergraph dynamical systems, the values of the nodes of each hyperedge are transformed by a possibly non-linear function $f_d$, and this update is projected back onto the nodes by summing over each update. 
In the spirit of this framework, we introduce a neural network-based learning approach for dynamics on hypergraphs. 
We call our framework Hypergraph Dynamics Graph Neural Network (\model).
Apart from minor technical adaptations, which are described in more detail in \cref{sec:hypergraphlearningmodel}, the main difference between \model~and the analytical formulation of our dynamics is that in \model, we approximate the update function $f_d$ using multi-layer perceptrons (MLPs).
We thus do not need to specify the functional form of the dynamics a priori, but can learn the dynamics entirely from observational data, as MLPs are universal function approximators that can approximate any function to arbitrary accuracy.

In order to estimate the effective (minimal) order of the dynamics, we consider \model s operating on MLPs with different input dimensions. 
For a \model{} of order $p_\text{model}$ only MLPs with up to $p_\text{model}$ input variables are used to model the $p$-dimensional functions $\phi^p_d$ in \cref{def:dynamical_order}. 
For a general multi-set $\mathfrak{s}$ of size $d-1$ we then have
\[
f_d(y_i,\mathfrak{s}) \approx \hat{f}_d(y_i,\mathfrak{s}) = \begin{cases} \text{MLP}_d^{d}(y_i,\mathfrak{s}) & \text{ if } p=d \\ \sum_{\mathfrak{v} \subseteq \mathfrak{s} : |\mathfrak{v}|  = p-1}\text{MLP}_d^{p}(y_i,\mathfrak{v}) & \text{ if } p<d
  \end{cases}
\]

We train our \model{} using datasets $\mathcal X=\{(x^{(1)}, \dot{x}^{(1)}),\dots,(x^{(S)}, \dot{x}^{(S)})\}$ consisting of pairs $(x^{(i)}, \dot{x}^{(i)})$ of state vectors $x^{(i)}$ and gradients $\dot{x}^{(i)}$ of the dynamics.
Our objective is for our model to output the state derivative vector $\dot{x}$, when the state vector $x$ is given as input:
\begin{equation}
  \dot{x} \approx \hat{M}(x) = \sum_{d=2}^k \; \sum_{ E \in \mathcal{E}_d: i \in E} \hat{f}_d (x_i, \multiset{x_j \mid j \in E, j \neq i}),
\end{equation}
where $\hat{f}_d$ describes the (estimated) update functions of our \model.
In particular, we use empirical risk minimization with a regularised $L_1$ loss function and fit the parameters $\theta_{p_{\text{model}}}$ of \model, which consist of the vector of all weights of the MLPs, by minimising the absolute difference of the prediction outputs and the provided target values of the training set $x$:
\begin{equation}
  \label{eq:loss_function}
  L(\mathcal X, \theta_{p_{\text{model}}}) = \sum_{i=1}^S \|\hat{M}(x^{(i)})-\dot{x}^{(i)}\|_1 + \lambda \|\theta_{p_{\text{model}}}\|_2
\end{equation}
The hyperparameter $\lambda$ of the $L_2$ penalty term is optimised through a hyperparameter search.

A technical detail we need to take care of in this context is that we must assign an arbitrary ordering to the values in the multiset $\mathfrak{s}$ (and thus also in the subsets $\mathfrak{v}$).
To alleviate this problem, we train the MLPs based on all possible orderings of the values in the multiset $\mathfrak{s}$ (or the subsets $\mathfrak{v}$) and take the mean to obtain the approximation of the $p$-ary function $\phi^p_d$ (see \Cref{sec:hypergraphlearningmodel}).

\subsection{Finding the effective model order}

Using the above outlined learning paradigm, we fit a series of \model s with increasing orders $p_{\text{model}}\leq k$ (i.e., we fit models of orders up to $k$).
Note that, if the chosen model order is less than the true dynamical order of the system, the \model{} will not fit well, since the MLPs of order $p_{\text{model}}<p_{\text{dyn}}$ cannot yield a good approximation to the nonlinear dynamics induced by the larger hyperedges. 
However, if the model order is equal to or greater than the dynamical order of the system, the dynamics will be learned accurately. 
As we take the topological order $k$ as an upper bound, the true (a priori) dynamical order of the dynamics may be larger than $p_{\text{min}}$.
However, if we find a model with an effective order that is smaller than the topological order, this means we have found a model order that approximates the true dynamics sufficiently well.
This may happen, e.g., if the dynamics are inherently reducible to a smaller order, i.e. their dynamical order is smaller than the topological order of the hypergraph.

Hence, to find the effective model order we need to find the model order that approximates the dynamics sufficiently well, while being as small as possible.
Depending on the application considered, there are multiple possible ways to select such an order.
For instance, we can select the order by manual inspection, or by setting a particular approximation threshold for the approximation quality of the dynamics, and choosing the smallest model order that achieves this threshold.

While this is not the main focus of our work here, we discuss in the following a simple scheme to automate this process of model selection.
Specifically, we define a model-corrected performance score to select the model with the lowest order $p_{\text{model}}$ that produces accurate results on a given dataset, as follows: 
\begin{align}
\label{eq:MC_perf}
\text{MC-perf}({p_{\text{model}}} | \mathcal X) =  \exp\left(-\frac{L(\mathcal X, \theta_{p_{\text{model}}})}{L_{\text{max}}}\right)\exp\left(-\frac{p_{\text{model}}}{k}\right)
\end{align}
where $L_{\text{max}}=\text{max}_{p_{\text{model}}}\left(L(\mathcal X,\theta_{p_{\text{model}}})\right)$.
We provide more details on the derivation of this model-corrected performance score in the Supplementary Materials in Section 2.
\section{Results}
In this section,  we show that we can indeed use our framework to learn the update function on the hyperedges from data. We then show that we can use these results to infer the effective order of the hypergraph dynamical system.

\subsection{Datasets}
We perform our experiments on synthetic Erd\H{o}s-Réyni hypergraphs and on a real hypergraph of high school students' contact patterns (SocioPatterns dataset). 
Both types of hypergraphs have topological order $k=4$.
On these hypergraphs, we simulate (higher-order variants of) four common linear and non-linear model: Kuramoto dynamics (synchronisation), SI dynamics (epidemic spread), multi-way consensus (MCM) dynamics (opinion dynamics), and linear diffusion. 
For each of these four dynamics we consider different variations of the dynamics, where we restrict the update functions $f_d$ to be a sum of at most $p$-ary update functions.
Details on the generation of the hypergraphs and the dynamics can be found in \Cref{sec:hypergraphlearningmodel}.

We look into two possible scenarios which may occur in the real world: In the first scenario (I) we observe the state vectors and their derivatives for particular time-points. 
In the second scenario (II), we observe a set of trajectories of the dynamics. 

For each of these settings, we create a corresponding dataset which we split into a training and test set. 
For the setting (I), each dataset consists of $500$ independently sampled random initial state vectors and the corresponding derivatives of the dynamics at that point (see \Cref{ssec:Datasets_methods}). 
This data can directly be used for training of the \model via empirical risk minimization (see \cref{eq:loss_function}).
For setting  (II), we generate a set of $25$ trajectories, whose initial state vectors are uniformly sampled (for details see \cref{ssec:Datasets_methods}). 
Starting from the random initial conditions, we simulate $100$ time steps with a time delta of $\Delta =0.01$ via a simple forward Euler scheme.
From these trajectories we then extract a training sequence for our model, by approximating the time-derivatives $\dot{x}$ via a simple forward time difference at each time point of the trajectories. 
We thus create a training set of $2500$ tuples of state vectors and associated derivatives which we can use to train our model.
We remark that in contrast to scenario (I) the training samples in scenario (II) are in fact not independent, as they are coupled over time via the governing equations of the dynamics.
To differentiate between the two scenarios we will refer to the training data in the first setting (I) as point-based training data, and the training data in the second setting (II) as trajectory-based training data.

\subsection{Numerical experiments: Learning the update functions}

\begin{figure*}[htb!]
  \centering
  \begin{subfigure}{0.45\textwidth}
    \includegraphics[width=\textwidth]{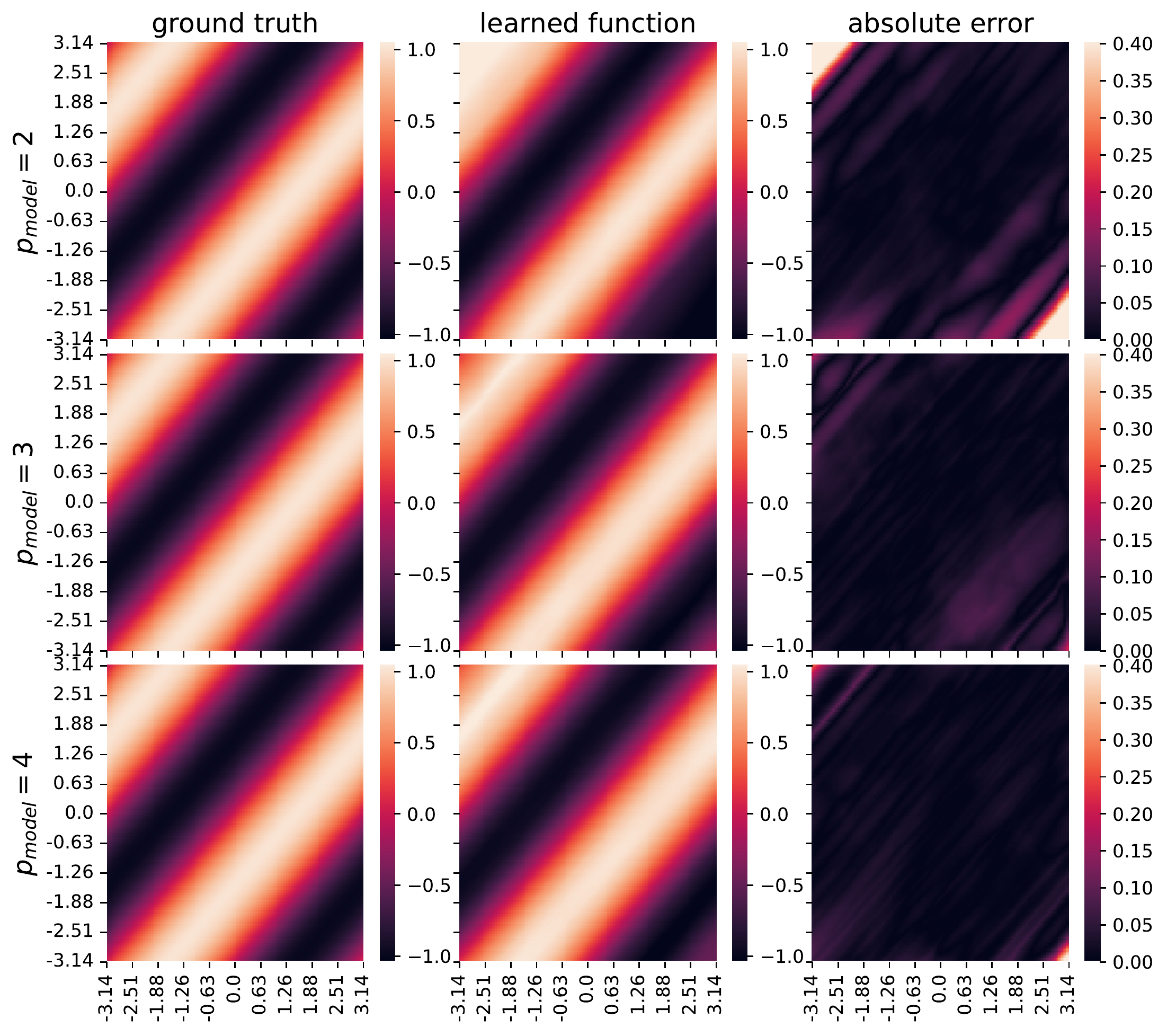}
    \subcaption{Learned Kuramoto update functions.}
  \end{subfigure}
  \begin{subfigure}{0.45\textwidth}
    \includegraphics[width=\textwidth]{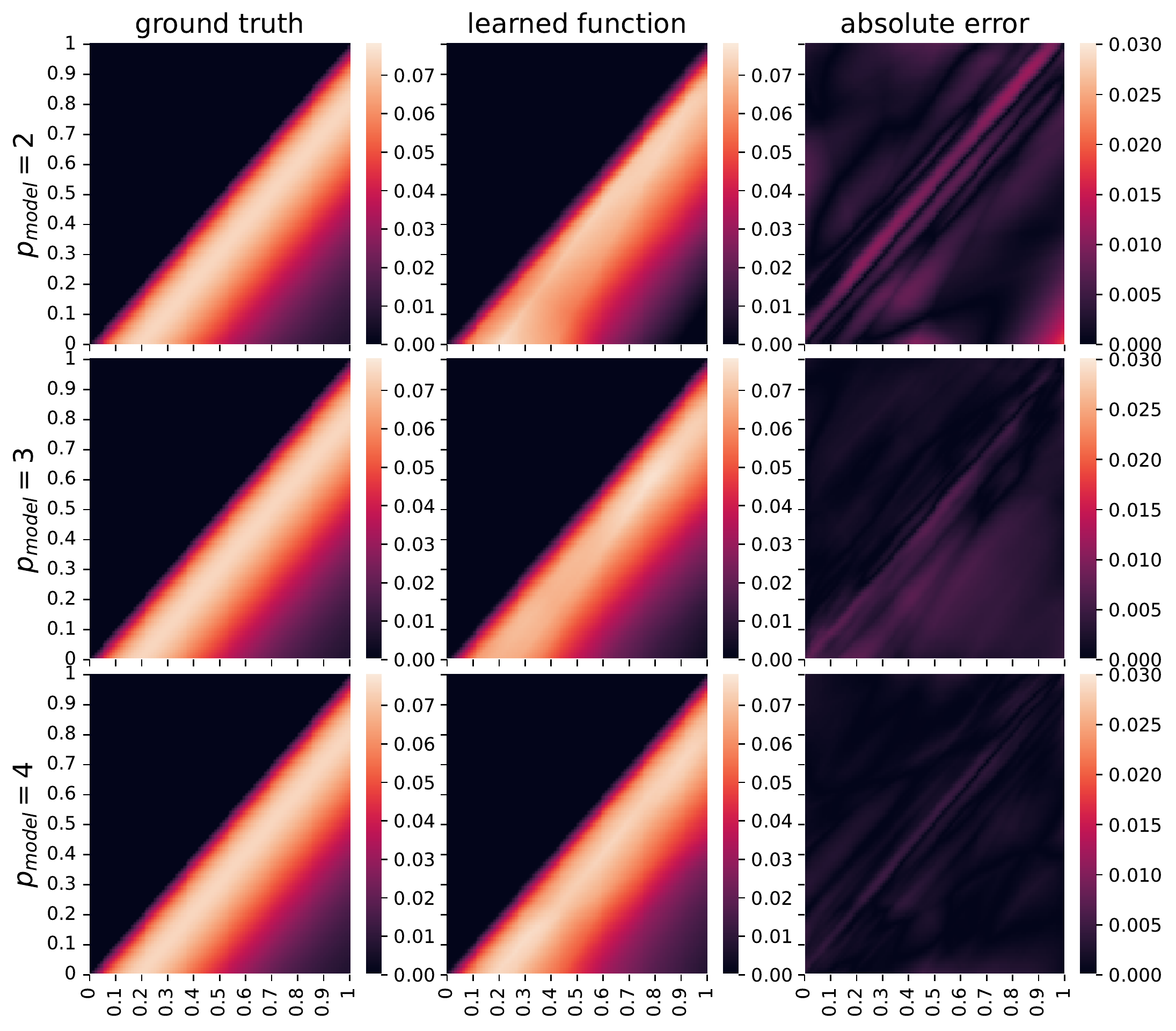}
    \subcaption{Learned MCM update functions.}
  \end{subfigure}
\caption[Learned update functions.]{\textbf{Learned update functions.} Here we see the ground truth (left columns) and learned update functions (middle columns) resulting from a \model{} of order $p_{\text{model}}\in\left\{2,3,4\right\}$ (top to bottom) trained on a synthetic derivative dataset of pairwise Kuramoto dynamics (a) and pairwise MCM dynamics (b) of order $p=p_{\text{dyn}}=2$. In particular, the x- and y values correspond to a range of two-dimensional inputs of the functions and we plot the corresponding function value. In the case when $p_{\text{model}}>2$, we only plot the output of the $\text{MLP}_d^{2}$ with input dimension 2. The ground truth update functions are shown in the left and the absolute errors of the approximations, which are given by the absolute difference of the individual function values of the ground truth and learned dynamics, are displayed in the right columns.  We observe a very good approximation to the ground truth dynamics for all model orders and both types of dynamics. This implies that our framework allows us to learn the update function of a dynamical system from data. As this is true for all model orders, we can additionally conclude that the framework can separate the effects of topology and dynamics, as it accurately captures the pairwise update functions even for higher model orders.}
\label{fig:edge-functions}
\end{figure*}

As a first step, we test whether a \model{} can accurately learn the dynamics from data. 
To this end, we first consider all dynamics restricted to dynamical order $p_{\text{dyn}}=2$ such that \model of any order should be sufficient to fit the dynamics.
We then train a \model{} with order $p_{\text{model}}\in\{2,3,4\}$ to see whether we can indeed learn the update function well.

In \Cref{fig:edge-functions} we see the true and learned update functions of a hypergraph neural network trained on a point-based training set of (a)~pairwise Kuramoto dynamics ($p_{\text{dyn}}=2$) and (b)~pairwise MCM dynamics ($p_{\text{dyn}}=2$).
We plot the true update functions as a function of the first and second input variable in the left columns. 
We show the learned dynamics for our \model s with model orders $p_{\text{model}}\in\left\{2,3,4\right\}$ (from top to bottom) in the middle columns. 
When $p_{\text{model}}>2$, we plot the output of the $\text{MLP}_d^{2}$ with input dimension 2 for simplicity.
The absolute error of the approximation, i.e., the absolute difference of the individual function values of the ground truth and learned dynamics, is shown in the right columns in \Cref{fig:edge-functions} (a) and (b).
As expected, we observe a good approximation of the ground truth dynamics for all \model~orders irrespective of the types of dynamics. 
In the Supplementary Materials in Figure 1, we show that for other dynamics we obtain equally good results.

\subsection{Numerical Experiments: Learning the hypergraph dynamical system}

\subsubsection{Synthetic Data}
\begin{figure*}
  \begin{subfigure}{\textwidth}
  \begin{subfigure}{0.69\textwidth}
    \includegraphics[width=\textwidth]{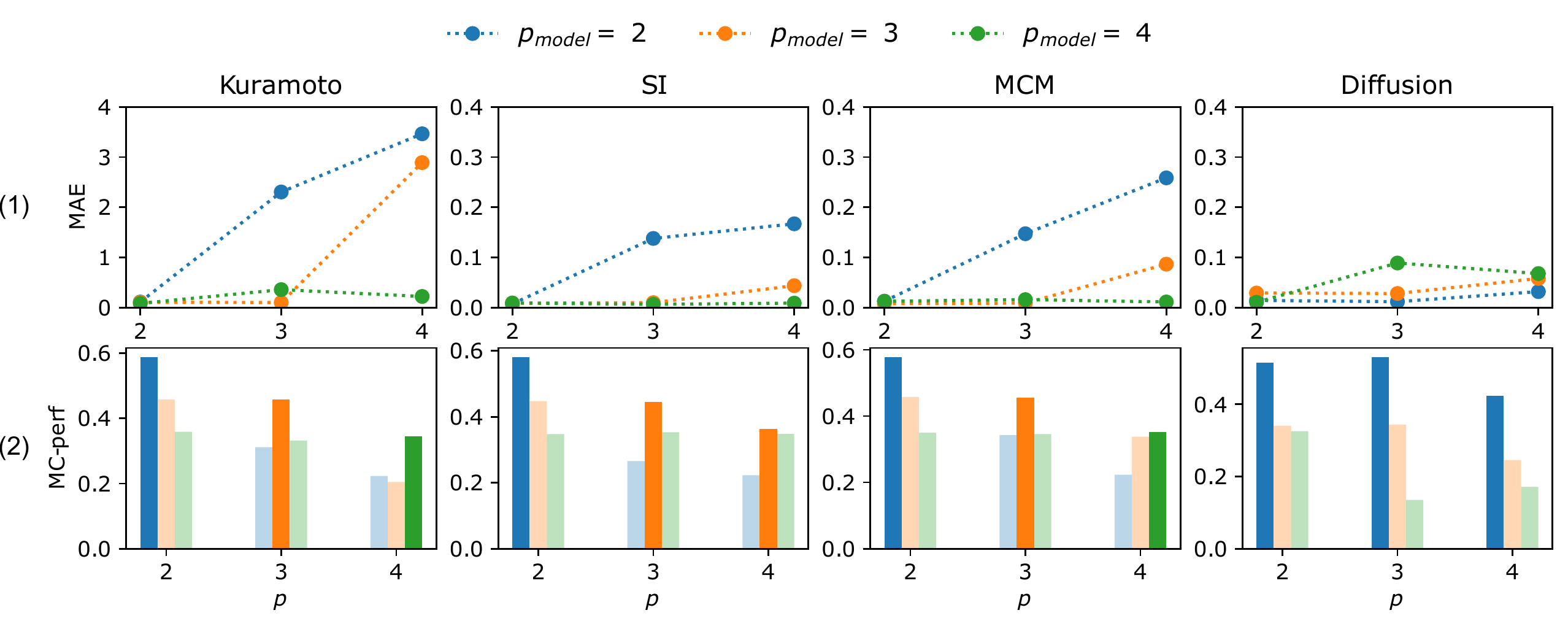}
    \subcaption{Learning results based on the derivative training set}
  \end{subfigure}
  \begin{subfigure}{0.3\textwidth}
    \includegraphics[width=\textwidth]{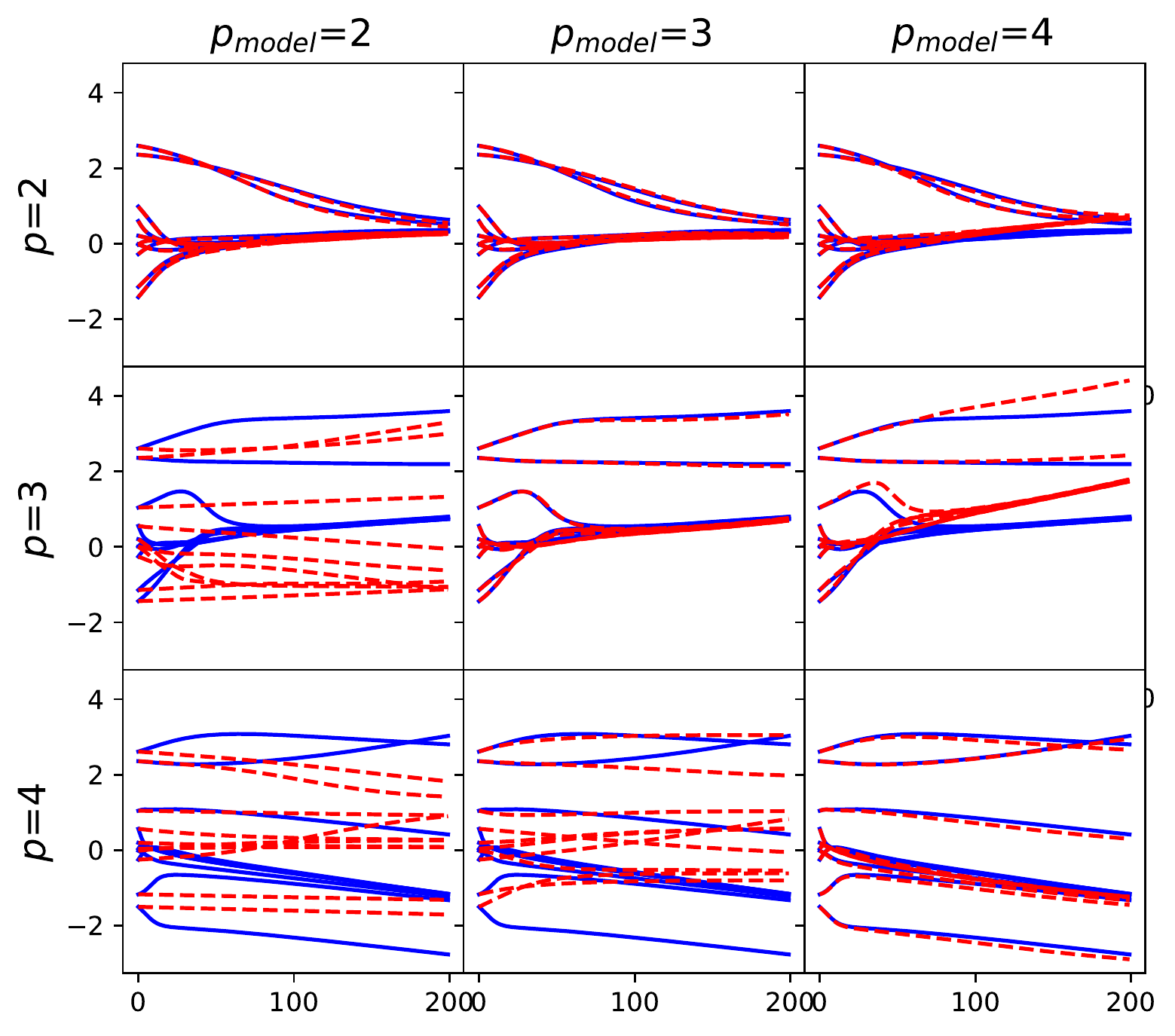}
    \subcaption{Kuramoto trajectories (der.)}
  \end{subfigure}
\end{subfigure}
\begin{subfigure}{\textwidth}
  \begin{subfigure}{0.69\textwidth}
    \includegraphics[width=\textwidth]{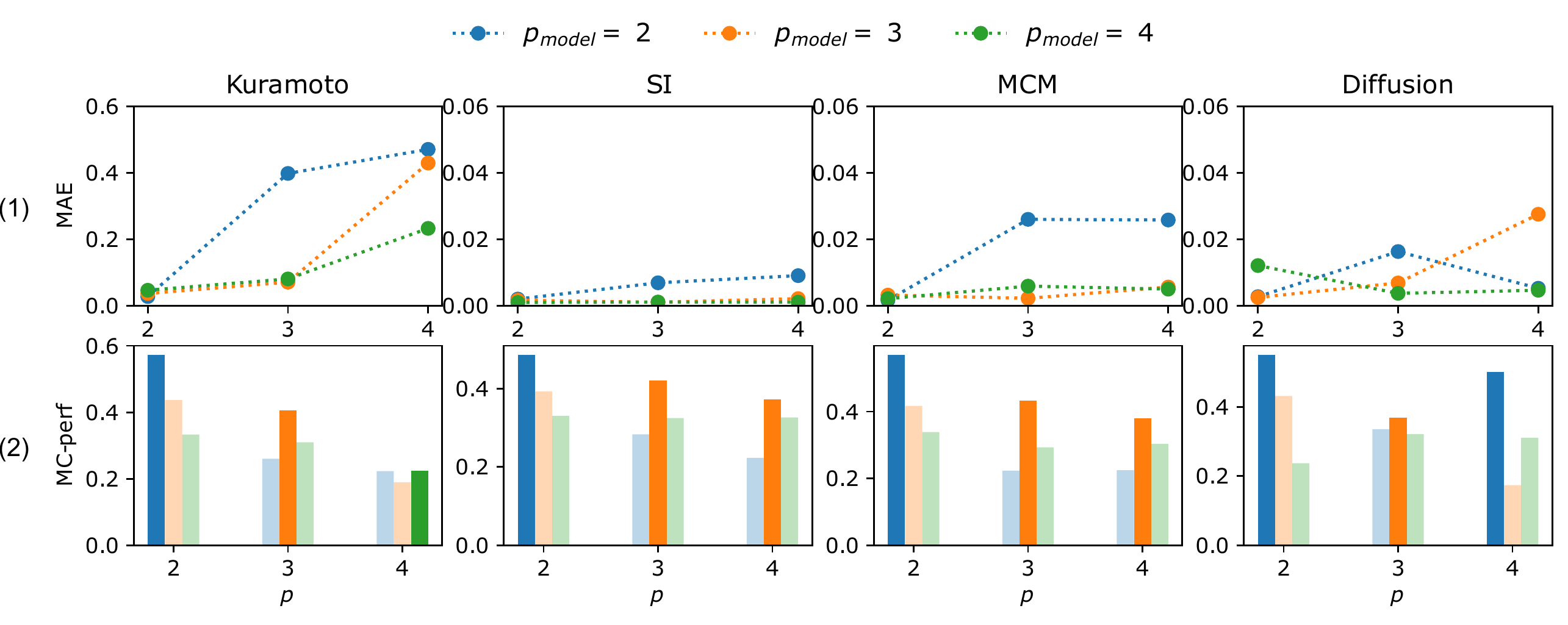}
    \subcaption{Learning results based on the trajectory training set}
  \end{subfigure}
  \begin{subfigure}{0.3\textwidth}
    \includegraphics[width=\textwidth]{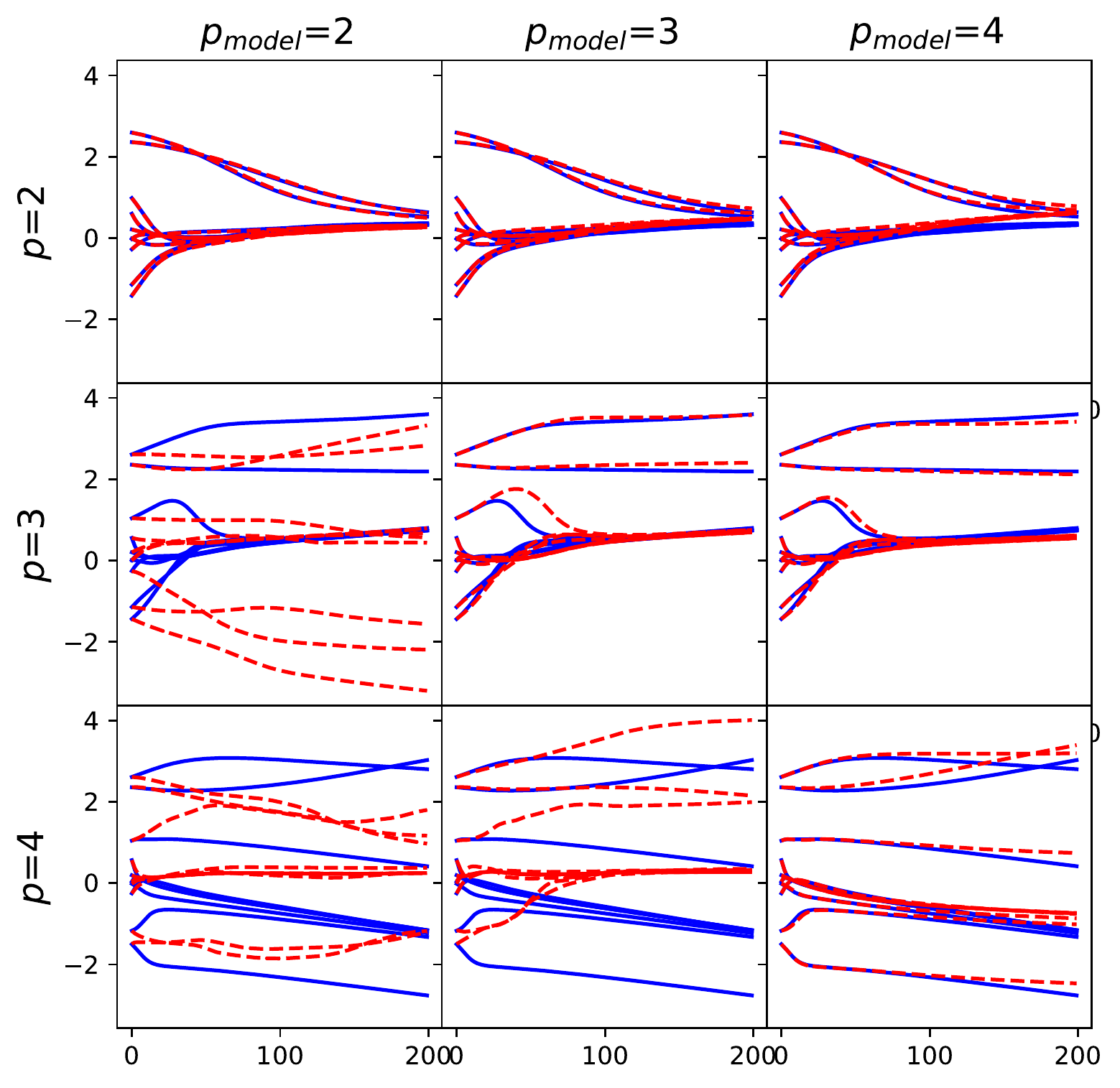}
    \subcaption{Kuramoto trajectories (traj.)}
  \end{subfigure}
\end{subfigure}
\caption[Experimental results on synthetic hypergraphs.]{\textbf{Experimental results on synthetic hypergraphs.} In this figure, we compare the learning results for the derivative ((a) and (b)) and the trajectory dataset ((c) and (d)) and show that we are indeed able to learn the effective order of a hypergraph dynamical system from data. 
    In (a.1), we display the mean absolute error (MAE) of cross-validation with 10 folds for all three models with $p_{\text{model}}\in\left\{2,3,4\right\}$ which were trained on the respective dynamics datasets (top row). 
    We can see that only if the model order is larger or equal to the effective order of the system (which is $p_{\text{min}}=p$ for Kuramoto, SI and MCM and $p_{\text{min}}=2$ for Diffusion), the dynamics can be learned by the \model. 
    In (a.2) we plot the model-corrected performance which is maximised for a model that optimises for both accuracy and model complexity. 
    We see that in almost all cases our performance score selects the model with $p_{\text{model}}=p_{\text{min}}$. 
    This shows that our method is indeed able to infer the effective order for the synthetic derivative training set. 
    Only in the case of SI dynamics, we can see that dynamics with effective order $p_{\text{min}}=4$ can already be roughly captured by an order $3$ model, which results in the selection of $p_{\text{model}}=3$ as the optimal order. 
    In (c), we evaluate the performance of our model for long-term trajectory predictions based on the trajectory training set. 
    In (c.1), we thus plot the MAE of ground truth and a predicted trajectory. 
    This evaluation thus includes accumulated errors. 
    We see that the results are similar to (a), whereas in this case the model-corrected performance score also suggests $p_{\text{min}}=3$ for a dynamics with  $p_{\text{dyn}}=4$ for MCM and SI, even though the model of order 4 leads to a lower MAE. 
    If a more accurate model is preferred, one could reweight the model-corrected performance score accordingly. 
    In the right column (b) and (d), we show an example of the long-term prediction of trajectories with $200$ time steps for the two datasets. 
    In particular, we plot ground truth Kuramoto trajectories (blue) with $p\in\left\{2,3,4\right\}$  (top to bottom) and the predicted trajectories given by a \model{} with $p_{\text{model}}\in\left\{2,3,4\right\}$  (left to right). 
    We can see that only models which are equal to or larger than the effective order of the system (which is given by $p_{\text{min}}=p$ for Kuramoto dynamics) can capture their long-term behaviour. 
    Moreover, the trajectory training set provides slightly better results. 
    This is because the trajectory dataset is biased towards the long-term behaviour of the dynamics. 
    As a result, it is more suitable for this type of long-term trajectory prediction. 
    Generally, the results confirm that our method is capable of learning the effective order, both for the derivative and trajectory training set and the resulting models can capture the long-term behaviour of the system.}
\label{fig:Results_synthetic}
\end{figure*}

We now demonstrate how we can use our framework to learn the effective order of an observed dynamics. 
Specifically, we examine hypergraph dynamical systems with Kuramoto, SI, MCM and Diffusion dynamics defined in terms of $p$-ary update functions with $p \in \left\{2,3,4\right\}$ (see \Cref{ssec:Datasets_methods}).
Note that, in general, $p$ does not have to be the dynamical order as we only ensure that there exists some decomposition of the dynamics into $p$-variate functions, but this decomposition does not have to be the minimal one.
In fact, for the diffusion dynamics we always have $p_{\text{dyn}}=2$ due to linearity. 
However, for the other update functions chosen here, the dynamical order is indeed always equal to the chosen decomposition, i.e., $p_\text{dyn}=p$ in case of the Kuramoto, SI and MCM dynamics, as we show in the Supplementary Materials in Section 1.
As the topological order of all considered hypergraphs is $k=4$, the effective order is thus equal to the dynamical order for all hypergraph dynamical systems. 

The results of the experiments on synthetic datasets are shown in \Cref{fig:Results_synthetic}, where we compare the learning results for the derivative datatset in \Cref{fig:Results_synthetic} (a)-(b) and the trajectory dataset in \Cref{fig:Results_synthetic} (c)-(d). 
While we always trained the model with the pointwise $L_1$-loss, as discussed in \cref{ssec:learning}, we evaluate the performance with two measures: (a.1) pointwise mean absolute error (MAE) (c.1) trajectory MAE (mean MAE over a trajectory) (see Supplementary Materials for details). Both MAEs are calculated on a cross-validation with $10$ folds.
Our results show that we are indeed able to learn the effective order from data. 
In~\Cref{fig:Results_synthetic} (a.1), we show the performance of the different models with $p_{\text{model}}\in\left\{2,3,4\right\}$, which is given by the lowest mean absolute error (MAE) of a cross-validation with $10$ folds. 

We see in~\Cref{fig:Results_synthetic} that the models, whose order is equal or larger than the effective order of the hypergraph dynamical system, clearly outperform the models with an order smaller than the effective order. 
As the dynamical order of the diffusion dynamics is $p_{\text{dyn}}=2$ for all hypergraph diffusion systems, all models can adequately approximate the dynamics in this case. 
We plot the model corrected performance in \Cref{fig:Results_synthetic} (a.2).
We see that our performance score almost always selects the model with $p_{\text{model}}=p_{\text{min}}$, which demonstrates that our method is able to infer the effective order for the synthetic training data set. 
In the case of SI dynamics, we find that the observed dynamics is also approximated well by an order $3$ model, which results in the selection of $p_{\text{min}}=3$ as effective order in this particular case.

In \Cref{fig:Results_synthetic}(c), we evaluate the performance of our model for long-term trajectory predictions based on the trajectory-based data set. 
In~\Cref{fig:Results_synthetic} (c.1), we thus plot the trajectory MAE of ground truth and predicted trajectories.
We see that the results are similar to (a).
However, in this case the model-corrected performance score is calculated by substituting $L_{MAE}$ into \cref{eq:MC_perf}, as described in the Supplementary Materials in Section 3. The results also suggests the selection of an effective order of $p_{\text{min}}=3$ for MCM and SI. An evaluation using the pointwise MAE which leads to similar results can be found in the Supplementary Materials in Figure 2. 

In~\Cref{fig:Results_synthetic} (b) and (d), we show an example of the long-term prediction of trajectories with $200$ time steps for the two datasets. Long-term prediction here means that we provide one initial condition and integrate over $200$ time steps with a time delta of $\Delta=0.01$ via a forward Euler scheme.
In particular, we plot ground truth Kuramoto trajectories (blue) with $p_\text{dyn}\in\left\{2,3,4\right\}$  (top to bottom) and the predicted trajectories given by a \model{} with $p_{\text{model}}\in\left\{2,3,4\right\}$ (left to right). 

We find that only a \model~with large enough order approximates the long-term behaviour of the dynamics well, as indicated also by the results in (a) and (c). 
Although the trajectory training set derived from a smaller amount of different hypergraph topologies (only $25$ different hypergraphs are considered instead of $500$ in the derivative dataset), it provides a better approximation of the trajectories (e.g., compare (b) and (d) for $p_{\text{model}}=4, p_{\text{dyn}}=3$). 

\begin{figure}[ht!]
  \centering
    \includegraphics[width=\columnwidth]{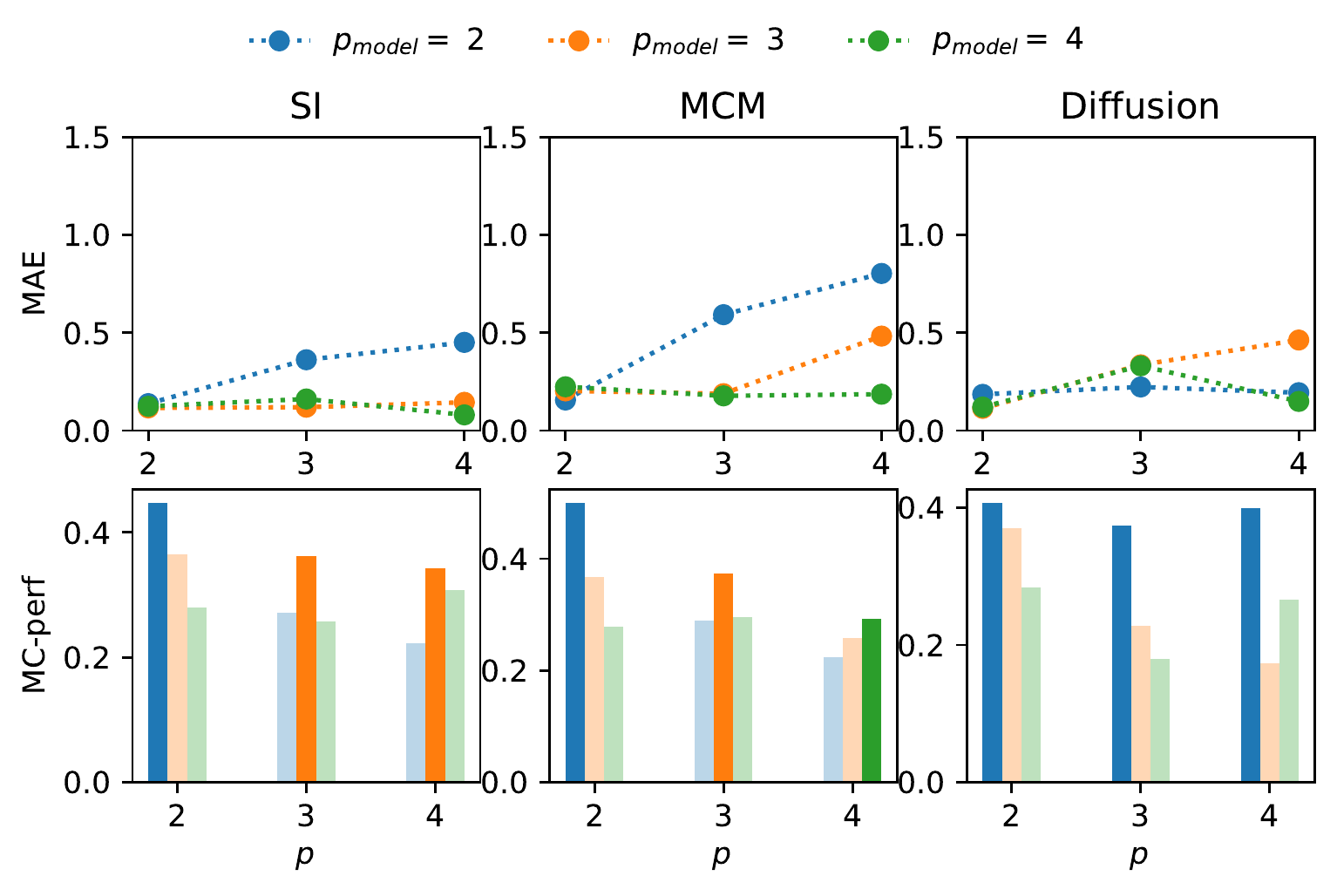}
  \caption[Experimental results on a real-world contact pattern hypergraph.]{\textbf{Experimental results on a real-world contact pattern hypergraph.} In this figure we display the results of experiments on three real-world datasets. In particular, we simulate epidemic spread (SI), opinion dynamics (MCM) and diffusion with order $p\in\left\{2,3,4\right\}$ as all of these dynamics can be expected to appear on a contact pattern hypergraph.  In (1), we again show the MAE of cross-validation with 10 folds for models with $p_{\text{model}}\in\left\{2,3,4\right\}$ which are trained on derivative datasets. As in the case of the synthetic datasets, we can see that only if the model order is larger or equal to the effective order, the trajectories can be learned by the hypergraph neural networks. This is undermined by the results of the model-corrected performance in (2). We see that the model order of the best model corresponds to the effective order of the system. Again, only in the
  case of SI dynamics, we can see that $p_{\text{dyn}}=4$ can already be roughly captured by a model with $p_{\text{model}}=3$. Overall, the results highlight the importance of carefully considering the type of dynamics when working with real-world hypergraph structures.}
  \label{fig:real_datasets}
\end{figure}

\subsubsection{Real-word datasets}
Similar to the results on the synthetic datasets, the results in \Cref{fig:real_datasets} show that using our method we can learn the effective order of dynamics unfolding on also on a real-world hypergraph of high-school student contacts.
Here, we consider the three interaction dynamics which would be expected on a contact dataset, epidemic spread (SI), opinion dynamics (MCM) and diffusion (e.g. diffusion of ideas or information). 
For diffusion, a linear dynamics, the inferred effective order is correctly inferred as $p_\text{min}=2$. 
For nonlinear dynamics such as epidemic spread and opinion formation, the best-performing model corresponds to an effective order $p_{\text{min}}=\left\{2,3,4\right\}$ rather than the order of the hypergraph.
This highlights the importance of carefully considering the type of dynamics when working with real-world hypergraph structures. 
Although this will not always hold, a lower-order hypergraph will be sufficient to capture the relevant dynamics in many cases.
Our method enables researchers to derive how much they can simplify their models from their specific datasets of interest.

\section{Discussion}

We presented a framework to infer a hypergraph dynamical system, trading off topological and dynamical complexity.
We used this framework to derive an effective hypergraph dynamical system for a given dynamics both analytically and in a data-driven way, based on observational data. 
In particular, using a neural network as a flexible way to approximate the local interaction dynamics, we were able to accurately learn the hypergraph dynamics and reduce the hypergraph order, while respecting the observed dynamics. 
In this context, finding an effective dynamical order that is smaller than the topological order of the hypergraph, indicates that we can ``prune down'' some hyperedges to be of smaller order as they do not play a role for the specific dynamics of interest. 
This is interesting from a point of view of model complexity reduction.
Moreover, the learned models are capable of predicting the long-term behaviour of dynamical systems. 

We believe that our methodological approach has a wide range of potential applications and provides a starting point for a research question that is currently somewhat understudied: namely, the trade-off between topological and dynamical complexity on higher-order domains such as hypergraphs, simplicial and cellular complexes.
Future research is needed to further investigate this connection between (hypergraph) topology, dynamics and effective order. 

\section{Methods}
\label{sec:hypergraphlearningmodel}
In this section, we formalise our hypergraph dynamical system model in more detail.
In particular, we consider a dynamics in terms of the topological structure and the local dynamics.
The topology of the system determines \textit{which} components of the system interact \textit{locally}.
The dynamics and specifically the update functions on the edges determine \textit{how} the components interact (update), and how these local interactions are \textit{globally} combined (projection) is again determined by the topology. 
\subsection{Hypergraph dynamical systems: Analytical framework}
Let $\mathcal{V}=\{1, \dots, N\}$ be a set of nodes in the system, represented by vertices. Let $x(t) \in \mathbb{R}^n$ be the vector of dynamical state variables of the nodes and $t$ the time. 
The hypergraph topology is given by a set of hyperedges $\mathcal{E}= \{ E_1, \dots, E_M \}$. 
For simplicity we group the hyperedges according to their size $d$ to obtain the whole hyperedge set $\mathcal{E}=\mathcal{E}_1 \cup \dots \cup \mathcal{E}_k$ where $\mathcal{E}_d=\{E_1^d, \dots, E^d_{M_d}\}$ and $E_\alpha^d$ is an arbitrary hyperedge with index $\alpha$ and size $d$ i.e. 
$|E_\alpha^d| = d$. 
The topological order of the hypergraph is defined by its maximal hyperedge size $k$. 
In the case of networks, we have that $d=2$ as the sets represent edges and thus only consist of two nodes $E^2_\alpha=\{i,j\}$. 
For $d>2$ the sets $E^d_\alpha=\{i,j, \dots, m\}$ represent hyperedges in which more than two nodes interact locally.

\begin{figure}
  \centering
  \begin{subfigure}{0.4\textwidth}
    \includegraphics[width=\textwidth]{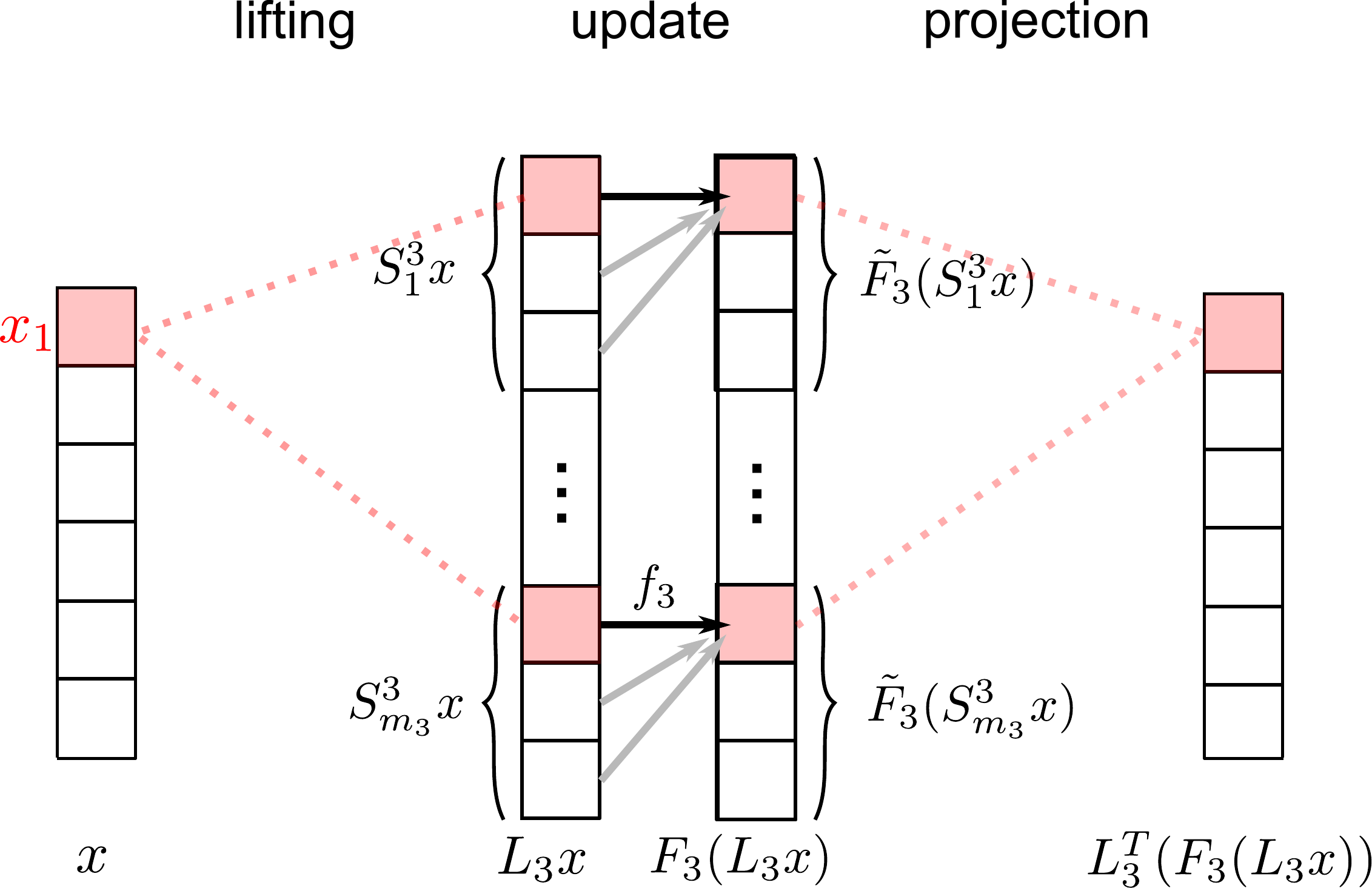}
  \end{subfigure}
  \caption[Hypergraph dynamics model.]{\textbf{Hypergraph dynamics model.} In this figure we show the update process of a single node according to the hyperedges of order $3$ in detail. Using the lifting operator $L_3$, the values of nodes that are part of hyperedges of size $3$ are collected. Then, a possibly non-linear function on these hyperedges is computed to obtain the update for a specific node, which is centred in these update functions. Subsequently, the obtained update values are projected back down to the node space by summing over each update.}
  \label{fig:1}
\end{figure}

\paragraph*{Topology}
For each $d$-hyperedge $E^d_\alpha =\{i,j,\dots,k\}$ we define its indicator matrix 
\begin{align}
    S^d_\alpha = \begin{bmatrix}
        e_i^\top\\
        e_j^\top\\
        \vdots\\
        e_k^\top
    \end{bmatrix}\in \mathbb{R}^{d\times n},
\end{align}
such that $S^d_\alpha x$ maps the node state vector $x$ to the node state vector $x_{E^d_\alpha}$ which only contains the state variables of the nodes included in hyperedge $E^d_\alpha$.  
We can then construct a (lifted) state vector by concatenating all the individual indicator vectors of the hyperedges by multiplying the original state vector $x$ with a matrix $L_d \in \mathbb{R}^{d|\mathcal{E}_d| \times n}$ defined as:
\begin{align}\label{eq:def_L}
  L_d = \begin{pmatrix} (S^d_1)^\top &\dots &(S^d_{m_d})^\top \end{pmatrix}^\top.
\end{align}
 The resulting vector corresponds to lifting the initial state vector into a higher-dimensional state space, hence we call the matrix a lifting matrix.
 The lifting matrix determines \textit{which} components of the system interact, represented by its non-zero entries (see~\Cref{fig:1}).

\paragraph*{Dynamics}
The (possibly non-linear) dynamics are then defined by the update operator $F:\mathbb{R}^{d|\mathcal{E}_d|} \rightarrow \mathbb{R}^{d|\mathcal{E}_d|}$. This operator consists of components $\tilde{F}:\mathbb{R}^{d} \rightarrow \mathbb{R}^{d}$ which defines the update for each hyperedge:
\begin{align}\label{eq:def_F}
  F_d(L_dx) = \begin{pmatrix} \tilde{F}_d(S_1^dx)^\top & \tilde{F}_d(S_2^dx)^\top & \dots & \tilde{F}_d(S_{m_d}^dx)^\top \end{pmatrix}^\top.
\end{align}
Even though the nodes are ordered, the update function is effectively a function on the interaction set and structured as:
\begin{align}\label{eq:def_blockwise_structure_F}
  \tilde{F}_d(y) = \begin{pmatrix}
    f_d(y_1, \multiset{y_1, ..., y_d}\backslash y_1) \\ 
    f_d(y_2, \multiset{y_1, ..., y_d}\backslash y_2) \\
    \vdots
    \\
    f_d(y_d, \multiset{y_1, ..., y_d}\backslash y_d) \\
  \end{pmatrix},
\end{align}
where we recall that the individual interaction functions $f_d$ are invariant to a permutation of their last $d-1$ arguments.
In other words, the function $f_d(x_i,\multiset{x_j(t) \mid j \in E^d_\alpha, j \neq i})$ computes the update for node $i$ resulting from the hyperedge $E_\alpha^d$. 

Following this generally non-linear update step, the updates are projected back onto the nodes via the transposed lifting operator, which simply sums up the updates on all hyperedges that a node is a part of. 
This procedure is displayed in \Cref{fig:1}. This results in the following global dynamics:
\begin{align}\label{eq:def_dynamics}
  \dot{x} = \sum_{d=2}^k L_d^\top F_d(L_dx)
\end{align}
For each node $i$ the dynamics can thus be written as
\begin{align}
\dot{x_i} &=  \left(\sum_{d=2}^k L_d^\top F(L_dx)\right)_i = \sum_{d=2}^k\left( \sum_{\alpha=1}^{M_d} \left(S_\alpha^\top \tilde{F}_d(S_\alpha^d x) \right)\right)_i \\&=  \sum_{d=2}^k \sum_{\alpha: i \in E_\alpha^d}   f_d \left(x_i, \multiset{x_j(t) \mid j \in E, j \neq i}\right).
\end{align}

Note that, since $F_d$ is equivariant under permutation of the ordering of the nodes in each hyperedge, the dynamics is \textit{globally} equivariant under permutation of the nodes. 
This restricts the dynamics that are possible. For example,
$$
f_d(y_1, \multiset{y_2, \cdots y_d}) = \beta y_1 + \gamma \sum_{i=1}^d y_i
$$ 
is the only possible dynamics in this framework that can be called linear. Still, many commonly used dynamical systems can be rewritten in our framework.

\subsection{Technical details of \model{}} 
In our analytical framework for hypergraph dynamical systems (\cref{eq:def_L,eq:def_F,eq:def_blockwise_structure_F}) the values of nodes of each hyperedge are collected by the lifting operator $L_d$, then transformed by a possibly non-linear function $f_d$ and this update is projected back down to the node space by summing over each update. 
As already stated in the main part of this work, we adapt this framework to introduce a neural network-based learning approach for dynamics on hypergraphs. 
We call our hypergraph dynamics learning model (\model). 

A technical issue in this context is that the input for the update functions in our analytical framework are \textit{sets} of state variable, which is a type of input Neural Networks cannot deal with. 
We thus derive equivalent formulations for \cref{eq:def_F,eq:def_blockwise_structure_F} which are consistent with the technical requirements of Neural Networks. 
Let $\Lambda(\mathfrak{s})$ be all possible orderings of a multi-set $\mathfrak{s}$. We can equivalently rewrite $f_d$ into ordered functions by using $f'_d(y_1, \cdots, y_d)=\frac{1}{d !}\sum f_d(y_1, \multiset{y_2,\cdots,y_d})$. Instead of \cref{eq:def_blockwise_structure_F}, in \model{} we define 
\begin{align}\label{eq:def_blockwise_structure_F_neural_network}
  \tilde{F'}^p_d(y) &= \begin{pmatrix}
    \sum_{\mathfrak{w} \in \Lambda(\multiset{y_1, ..., y_d}\backslash y_1)}f'_d(y_1, \mathfrak{w}) \\ 
    \sum_{\mathfrak{w} \in \Lambda(\multiset{y_1, ..., y_d}\backslash y_2)}f'_d(y_2,\mathfrak{w}) \\
    \vdots
    \\
    \sum_{\mathfrak{w} \in \Lambda(\multiset{y_1, ..., y_d}\backslash y_d)}f'_d(y_d, \mathfrak{w}) \\
  \end{pmatrix} \\&= \begin{pmatrix}
    \sum_{\mathfrak{v} \subseteq \multiset{y_1, ..., y_d}\backslash y_1 : |\mathfrak{v}|+1 =p }\sum_{\mathfrak{w} \in \Lambda(\mathfrak{v})}\phi^p_d(y_1, \mathfrak{w}) \\ 
    \sum_{\mathfrak{v} \subseteq \multiset{y_1, ..., y_d}\backslash y_2 : |\mathfrak{v}|+1 =p}\sum_{\mathfrak{w} \in \Lambda(\mathfrak{v})}\phi^p_d(y_1, \mathfrak{w}) \\
  \vdots
  \\
  \sum_{\mathfrak{v} \subseteq \multiset{y_1, ..., y_d}\backslash y_d : |\mathfrak{v}|+1 =p}\sum_{\mathfrak{w} \in \Lambda(\mathfrak{v})}\phi^p_d(y_1, \mathfrak{w})\\
\end{pmatrix}
\end{align}
and the global update function in \cref{eq:def_F} is then redefined as $F'_d$ by substituting in $\tilde{F'}^p_d$ instead of $\tilde{F}_d$ accordingly. 

This substitution results in the same dynamics as long as $f_d$ is decomposable in $p$-dimensional functions as in \cref{def:dynamical_order}. 
The model then still computes the global updates, i.e. the learned approximation to the dynamics given in the data as:
\begin{align}\label{eq:def_dynamics_neural_network}
  \dot{x} = \sum_{d=2}^k L_d^\top F'_d(L_dx)
\end{align}
Note that this comes at a computational price. 
To compute the updates for each node, all $d \!$ permutations are computed --- which limits the order of the system this framework can be applied to.

\subsection{Datasets and Dynamics considered in Numerical Experiments}
\label{ssec:Datasets_methods}
We perform our experiments on datasets of dynamics on synthetic (Erd\H{o}s-Réyni hypergraphs) and real-world topologies (contact pattern dataset of high school students). We assume that the graph topology is given in the form of an adjacency list from which we extract the lifting operator.  Based on this topology we simulate a specific dynamics of interest. 

In this work, we specifically focus on four common linear and non-linear dynamics in network science: Kuramoto dynamics (Synchronisation), SI dynamics (epidemic spread), Multi-way Consensus Model (MCM, opinion dynamics) and Diffusion. We define these dynamics here for general order $p$ in the notation of \cref{def:dynamical_order}:
\paragraph{Synchronisation (Kuramoto oscillator dynamics)}
The Kuramoto model~\cite{kuramoto_chemical_2003} has been applied to various synchronization phenomena of phase oscillators~\cite{Stankovski-2017}, ranging from power networks~\cite{dorfler_synchronization_2010} to brain activity~\cite{cabral_exploring_2014} and several works have been working on its generalisation to hypergraphs~\cite{adhikari_synchronization_2023}. The nodal state $x_i$ corresponds to the phase of oscillator $i$ and the (hyper)edges represent the couplings between the oscillators. 
\begin{align*}
  \phi^p_d(y_i, \multiset{y_1, \cdots, y_{p}}\backslash y_i)= \sin\left(\sum^p_{j =1} (y_j - y_i)\right)
\end{align*}
\paragraph{Epidemic Spread (Susceptible-Infected Model)}
The spreading of an infectious disease can be described by the simple susceptible–infected model (SI). The nodal state $x_i$ equals the infection probability of node $i$. The (hyper)edges represent infection rates between people.
\begin{align*}
  \phi^p_d(y_i, \multiset{y_1, \cdots, y_{p}}\backslash y_i) =(1-y_i)\prod^p_{j =1, j \neq i} y_j
\end{align*}
\paragraph{Opinion dynamics with reinforcement (Multi-way Consensus Model)}
The Multi-way Consensus Model (MCM)~\cite{neuhauser_multibody_2020,neuhauser_opinion_2021,sahasrabuddhe_modelling_2020,neuhauser_consensus_2022} models opinion formation with group effects. These effects are captured by a non-linear function which scales the consensus term of the model. Depending on its form, the scaling function can capture reinforcement effects of the members of a hyperedge and we specifically choose the model facet MCMI here, which models homophily:
\begin{align*}
  \phi^p_d(y_i, \multiset{y_1, \cdots, y_{p}}\backslash y_i) = \exp\left(-\left(\frac{\sum_{j =1}^py_j}{d}-y_i\right)\right)\sum_{j =1}^p (y_j - y_i)
\end{align*}
\paragraph{Linear consensus dynamics (Diffusion)}
The exchange of information between autonomous agents who seek some form of cooperation can be captured by this simple consensus protocol~\cite{somarakis_simple_2015}. The links hereby describe the nodes' influences on each other. Applications range from the spread of information in a social network to optimal control.
\begin{align*}
  \phi^p_d(y_i, \multiset{y_1, \cdots, y_{p}}\backslash y_i) = \sum_{ j =1}^p(y_j - y_i)
\end{align*}

\paragraph{Numerical Integration of dynamics}
We simulate these dynamics for different, random initialisations and different order $p\in\left\{2,3,4\right\}$ to obtain the input/output pairs: synchronisation (Kuramoto oscillator dynamics) with a uniform initialisation in $\left[-\pi,\pi\right]$, epidemic spread (SI) with a uniform initialisation in $\left[0,1\right]$, opinion dynamics (MCM) with a skewed initialisation in $\left[0,1\right]$ (as the nonlinear reinforcing effects mainly appear for skewed distributions) and spreading (diffusion) with a uniform initialisation in $\left[-1,1\right]$. The datasets then consist of observed samples of dynamics with different orders $p$ on the given hypergraph topology. 

\paragraph{Synthetic and real-world hypergraphs}
For our synthetically generated hypergraphs we used a set of 500 Erd\H{o}s-Réyni hypergraphs with 20 nodes. Any $2$ edge is created with probability $0.01$, any $3$-edge with probability $0.001$ and every $4$-edge with probability $0.001$. 
The hypergraph is generated by the xgi package according to \cite{dewar_subgraphs_2016}.

As real-world hypergraph example, we use a static version of a temporal higher-order dataset constructed from interactions recorded by wearable sensors worn by students at a high school~\cite{chodrow2021hypergraph, Mastrandrea-2015-contact}. 
The hypergraph consists of $327$ nodes and $7818$ hyperedges, with an average size of $2.3$ nodes and a maximum size of $5$. 
However, as the dataset only contains $7$ edges of size $5$, we reduced the hypergraph to consider hyperedges up to size $4$, resulting in a hypergraph of order $4$. 
The resulting dataset includes $222$ $4$-edges, $2091$ $3$-edges, and $5498$ $2$-edges. 
On this hypergraph, we simulate the three social dynamics which can occur on a contact dataset -- epidemic spread (SI), opinion dynamics with peer pressure (MCM), and diffusion.

\section{Data availability}
The SocioPatterns data are available at \url{https://www.cs.cornell.edu/~arb/data/#hyperlabels.
}

\section{Code availability}
The code will be made publicly available.

\section{Acknowledgements}
L.N., M.S. and M.T.S. acknowledge funding by the Ministry of Culture and Science (MKW) of the German State of North Rhine- Westphalia ("NRW Rückkehrprogramm") and under the Excellence Strategy of the Federal Government and the Länder. 

\section{Competing interests}
The authors declare that they have no competing interests.

\bibliography{bib}
\end{document}